\documentclass[10pt, a4paper]{article}
\usepackage{lrec}
\usepackage{multibib}
\newcites{languageresource}{Language Resources}
\usepackage{graphicx}
\usepackage{tabularx}
\usepackage{soul}
\usepackage{xspace}
\usepackage{amsmath}

\usepackage{booktabs}
\usepackage{multirow}
\usepackage{gb4e}
\noautomath

\usepackage{epstopdf}

\usepackage{hyperref}
\usepackage{xstring}
\usepackage[utf8]{inputenc}
\usepackage{tikz-dependency}

\newcommand{\smallsc}[1]{\scriptsize\textsc{#1}}
\newcommand{\F}{$\text{F}_1$\xspace}

\title{\textbf{MultiBooked: A Corpus of Basque and Catalan Hotel Reviews \\
               Annotated for Aspect-level Sentiment Classification.}}

\name{Jeremy Barnes, Patrik Lambert, Toni Badia}

\address{Pompeu Fabra University, 
          Barcelona, Spain\\
         \{firstname.lastname\}@upf.edu}

\abstract{
While sentiment analysis has become an established field in the NLP community, research into languages other than English has been hindered by the lack of resources. Although much research in multi-lingual and cross-lingual sentiment analysis has focused on unsupervised or semi-supervised approaches, these still require a large number of resources and do not reach the performance of supervised approaches. With this in mind, we introduce two datasets for supervised aspect-level sentiment analysis in Basque and Catalan, both of which are under-resourced languages. We provide high-quality annotations and benchmarks with the hope that they will be useful to the growing community of researchers working on these languages. 
 \\ \newline  }

\begin{document}

\maketitleabstract
\section{Introduction}

Sentiment analysis has become an established field with a number of subfields (aspect-level sentiment analysis, social media sentiment analysis, cross-lingual sentiment analysis), all of which require some kind of annotated resource, either to train a machine-learning based classifier or to test the performance of proposed approaches. 

Although much research into multi-lingual and cross-lingual sentiment analysis has focused on unsupervised or semi-supervised approaches \cite{Balamurali2012,Perezrosas2012,Gao2015}, these techniques still require certain resources (linked wordnets, seed lexicon) and do not generally reach the performance of supervised approaches. 

In English the state-of-the-art for binary sentiment analysis often reaches nearly 90 percent accuracy \cite{Tai2015a,Kim2014,Irsoy2014a}, but for other languages there is a marked drop in accuracy. This is mainly due to the lack of annotations and resources in these languages. This is especially true of corpora annotated at aspect-level. Unlike document- or tweet-level annotation, aspect-level annotation requires a large amount of effort from the annotators, which further reduces the likelihood of finding an aspect-level sentiment corpus in under-resourced languages. We are, however, aware of one corpus annotated for aspects in German \cite{Klinger2014a}, although German is not a particularly low-resource language. 

The movement towards multi-lingual datasets for sentiment analysis is important because many languages offer different challenges, such as complex morphology or highly productive word formation, which can not be overcome by focusing only on English data.

The novelty of this work lies in creating corpora which cover both Basque and Catalan languages and are annotated in such a way that they are compatible with similarly compiled corpora available in a number of languages \cite{Agerri2013}. This allows for further research into cross-lingual sentiment analysis, as well as introducing the first resource for aspect-level sentiment analysis in Catalan and Basque. The corpus is available at \url{http://hdl.handle.net/10230/33928} or \url{https://jbarnesspain.github.io/resources/}.

\section{Related Work}
In English there are many datasets available for document- and sentence-level sentiment analysis across different domains and at different levels of annotation \cite{Pang2002,HuandLiu2004,Blitzer2007,Socher2013b,Nakov2013}. These resources have been built up over a period of more than a decade and are currently necessary to achieve state-of-the-art performance.

Corpora annotated at fine-grained levels (opinion- or aspect-level) require more effort from annotators, but are able to capture information which is not present at document- or sentence-level, such as nested opinions or differing polarities of different aspects of a single entity. In English, the MPQA corpus \cite{Wiebe2005} has been widely used in fine-grained opinion research. More recently, a number of SemEval tasks have concentrated on aspect-level sentiment analysis \cite{Pontiki2014,Pontiki2015,Pontiki2016}.

The Iberian peninsula contains two official languages (Portuguese and Spanish), as well as three co-official languages (Basque, Catalan, and Galician) and several smaller languages (Aragonese, Gascon). The two official languages do have available resources for sentiment at tweet-level \cite{TASS2013,Arruda2015}, as well as at aspect-level \cite{Agerri2013,TASS2014,Almeida2015}. The co-official languages, however, have almost none. The authors are aware of a small discourse-related sentiment corpus available in Basque \cite{Alkorta2015}, as well as a stance corpus in Catalan \cite{Bosco2016}. These resources, however, are limited in size and scope.

\section{Data Collection}
In order to improve the lack of data in low-resource languages, we introduce two aspect-level sentiment datasets to the community, available for Catalan and Basque. To collect suitable corpora, we crawl hotel reviews from \url{www.booking.com}. Booking.com allows you to search for reviews in Catalan, but it does not include Basque. Therefore, for Basque we crawled reviews from a number of other websites that allow users to comment on their stay\footnote{We took reviews from a total of 35 different websites, including \url{www.airbnb.com}, 	\url{www.atrapalo.com}, \url{www.nekatur.net}, \url{www.rentalia.es}, \url{www.toprural.es}, and \url{www.tripadvisor.com}.} 

Many of the reviews that we found through crawling are either 1) in Spanish, 2) include a mix of Spanish and the target language, or 3) do not contain any sentiment phrases. Therefore, we use a simple language identification method\footnote{We use the count of stopwords to predict the probability that a review is written in Spanish, Catalan, or Basque.} in order to remove any Spanish or mixed reviews and also remove any reviews that are shorter than 7 tokens. This finally gave us a total of 568 reviews in Catalan and 343 reviews in Basque, collected from November 2015 to January 2016.

We preprocess them through a very light normalization, after which we perform tokenization, pos-tagging and lemmatization using Ixa-pipes \citelanguageresource{Agerri2014}.

Our final documents are in KAF/NAF format \cite{Bosma2009,Fokkens2014}. This is a stand-off xml format originally from the Kyoto project \cite{Bosma2009} and allows us to enrich our documents with many layers of linguistic information, such as the pos tag of a word, its lemma, whether it is a polar word, and if so, if it has an opinion holder or target. The advantage of this format is that we do not have to change the original text in any way.

\section{Annotation}

For annotation, we adopt the approach taken in the OpeNER project \cite{Agerri2013}, where annotators are free to choose both the span and label for any part of the text.

\subsection{Guidelines}
\label{guidelines}

In the OpeNER annotation scheme\footnote{\url{http://www.opener-project.eu/}} (see Table \ref{table:guidelines} for a short summary), an annotator reads a review and must first decide if there is any positive or negative attitudes in the sentence. If there are, they then decide if the sentence is on topic. Since these reviews are about hotels, we constrain the opinion targets and opinion expressions to those that deal with aspects of the hotel. Annotators should annotate the span of text which refers to:

\begin{itemize}
\item \textbf{opinion holders},
\item \textbf{opinion targets},
\item and \textbf{opinion expressions}.
\end{itemize}  

If any opinion expression is found, the annotators must then also determine the polarity of the expression, which can be \textsc{strong negative, negative, positive,} or \textsc{strong positive}. As the opinion holder and targets are often implicit, we only require that each review has at least one annotated opinion expression.

\begin{table}[htp!]
\centering
\begin{tabular}{l|c}
\toprule
Is there a positive / negative attitude? & yes/no \\
Is the sentence on topic ? & yes/no \\
Is it to the point? & yes/no \\
\\
IF YES TO ALL, ANNOTATE: &\\
What is the span of the expression? & choose span \\
Is the expression positive or negative? & choose \\
Is the expression strong? & choose \\
\\
Is there an explicit target? & yes/no\\
If yes, what is the span? & choose span \\
\\
Is there an explicit opinion holder & yes/no \\
If yes, what is the span? & choose span\\

\bottomrule
\end{tabular}
\caption{Simplified annotation guidelines.}
\label{table:guidelines}
\end{table}

For the strong positive and strong negative labels, annotators must use clues such as adverbial modifiers ('very bad'), inherently strong adjectives ('horrible'), and any use of capitalization, repetition, or punctuation ('BAAAAD!!!!!') in order to decide between the default polarity and the strong version.

\subsection{Process}

We used the KafAnnotator Tool \cite{Agerri2013} to annotate each review. This tool allows the user to select a span of tokens and to annotate them as any of the four labels mentioned in Section \ref{guidelines}. 

The annotation of each corpus was performed in three phases: first, each annotator annotated a small number of reviews (20-50), after which they compared annotations and discussed any differences. Second, the annotators annotated half of the remaining reviews and met again to discuss any new differences. Finally, they annotated the remaining reviews. For cases of conflict after the final iteration, a third annotator decided between the two.

The final Catalan corpus contains 567 annotated reviews and the final Basque corpus 343. 

\subsection{Dataset Characteristics}

The reviews are typical hotel reviews, which often mention various aspects of the hotel or experience and the polarity towards these aspects. An example is shown in Example

\begin{figure}
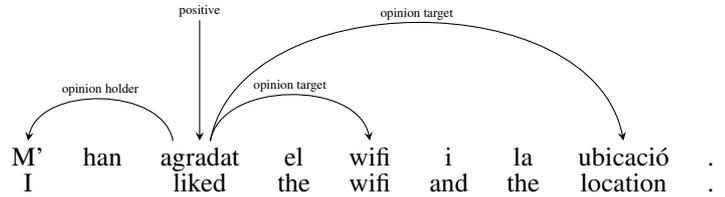

\begin{dependency}[theme = simple]
   \begin{deptext}[column sep=1em]
      M' \& han \& agradat \& el  \& wifi \& i \& la \& ubicació \& .\\
      I \& \& liked \& the \& wifi \& and \& the \& location \&. \\
   \end{deptext}
   \deproot{3}{positive}
   \depedge[edge start x offset=-6pt]{3}{1}{opinion holder}
   \depedge{3}{5}{opinion target}
   \depedge{3}{8}{opinion target}
\end{dependency}
\label{deptree}
\caption{An opinion annotation following the annotation scheme detailed in Section \ref{guidelines}.}
\end{figure}

Statistics for the two corpora are shown in Table \ref{table:stats}.

\begin{table}[htp!]
\centering
\begin{tabular}{lcc}
\toprule
& \multicolumn{1}{c}{Catalan} & \multicolumn{1}{c}{Basque}\\
\cmidrule(lr){1-1}\cmidrule(lr){2-2}\cmidrule{3-3}
 \textit{Number of Reviews} & 567 & 343 \\
 \textit{Average length in tokens} & 45&  46.9\\
 \textit{Number of Targets}   & 2762  & 1775 \\
 \textit{Number of Expressions} & 2346  & 2328  \\
 \textit{Number of Holders}     & 236   & 296 \\
\bottomrule
\end{tabular}
\caption{Corpus Statistics}
\label{table:stats}
\end{table}

\subsection{Agreement Scores}

Common metrics for determining inter-annotator agreement, e.g. Cohen's Kappa \cite{Cohen1960} or Fleiss' Kappa \cite{Fleiss1971}, can not be applied when annotating sequences, as the annotators are free to choose which parts of a sequence to include. Therefore, we use the \textit{agr} metric \cite{Wiebe2005}, which is defined as:

\begin{equation} \label{eq:1}
\textrm{agr}(a||b) = \frac{|A \textrm{ matching }B|}{|A|}
\end{equation}

where $a$ and $b$ are annotators and $A$ and $B$ are the set of annotations for each annotator. If we consider $a$ to be the gold standard, $agr$ corresponds to the recall of the system, and precision if $b$ is the gold standard. For each pair of annotations, we report the average of the $agr$ metric with both annotators as the temporary gold standard, 

\begin{equation} \label{eq:2}
\textrm{AvgAgr}(a,b) = \frac{1}{2} \big[ \textrm{agr}(a||b) + \textrm{agr}(b||a) \big]
\end{equation}

Perfect agreement, therefore, is 1.0 and no agreement whatsoever is 0.0. Similar annotation projects \cite{Wiebe2005} report $AvgAgr$ scores that range between 0.6 and 0.8 in general.

For polarity, we assign integers to each label (Strong Negative: 0, Negative: 1, Positive: 2, Strong Positive: 3). For each sentence of length $n$, we take the mean squared error (MSE),

\begin{equation} \label{eq:3}
\textrm{Mean Squared Error} = \frac{1}{n}\sum_{i=1}^{n}(A - B)^{2}
\end{equation}

where $A$ and $B$ are the sets of annotations for the sentence in question. This approach punishes larger discrepancies in polarity more than small discrepancies, i.e. if annotator 1 decides an opinion expression is \textsc{strong negative} and annotator two that the same expression is \textsc{positive}, this will be reflected in a larger MSE score than if annotator 2 had chosen \textsc{negative}. Perfect agreement between annotators would lead to a MSE of 0.0, with the maximum depending on the length of the phrase. For a phrase of ten words, the worst MSE possible (assuming annotator 1 labeled all words \textsc{strong positive} and annotator 2 labeled them \textsc{strong negative}) would be a 9.0. We take the mean of all the MSE scores in the corpus.

 Inter-annotator agreement is reported in Table \ref{table:IAA}.

\begin{table}[htp!]
\centering
\begin{tabular}{lcc}
\toprule
& \multicolumn{1}{c}{Catalan} & \multicolumn{1}{c}{Basque}\\
\cmidrule(lr){1-1}\cmidrule(lr){2-2}\cmidrule{3-3}
 \textit{Number of Reviews} & 567 & 343 \\
 \textit{Targets}   & .767  & .739 \\
 \textit{Expressions} & .716  & .714  \\
 \textit{Holders}     & .121   & .259 \\
 \textit{Polarity (MSE)}  & 1.53  & 2.7 \\
\bottomrule
\end{tabular}
\caption{Inter-annotator agreement scores. \textit{AvgAgr} score is reported for targets, expressions and holders and averaged mean squared error is reported for polarity. }
\label{table:IAA}
\end{table}

The inter-annotator agreement for target and expressions is high and in line with previous annotation efforts \cite{Wiebe2005}, given the fact that annotators could choose any span for these labels and were not limited to the number of annotations they could make. This reflects the clarity of the guidelines used to guide the annotation process. 

The agreement score for opinion holders is somewhat lower and stems from the fact that there were relatively few instances of explicit opinion holders. Additionally, Catalan and Basque both have agreement features for verbs, which could be considered an implicit mention of the opinion holder. This is not always clear, however. Finally, the mean squared error of the polarity scores shows that annotators generally agree on where and which polarity score should be given. Again, the mean squared error in this annotation scheme requires both annotators to choose the same span and the same polarity to achieve perfect agreement.

\section{Difficult Examples}

During annotation, there were certain sentences which presented a great deal of problems for the annotators. Many of these are difficult because of 1) \textbf{nested opinions}, 2) \textbf{implicit opinions reported only through the presence or absence of certain aspects}, or 3) \textbf{the difficulty to identify the span of an expression}. Here, we give examples of each difficulty and detail how these were resolved during the annotation process.

\begin{small}
\begin{exe}
\ex \label{nested} \gll Hotela bikaina zen , nahiz eta bertako langileak ez bereziki jatorrak izan. \\
Hotel\smallsc{.abs.sg} great\smallsc{.abs.sg} be , although {} there.from workers.\smallsc{abs.pl} not particularly friendly\smallsc{.abs.pl} were \\
\trans `The hotel was great, although the workers there were not particularly friendly.'
\end{exe}
\end{small}

In the Basque sentence in Example \ref{nested}, we can see that there are two distinct levels of aspects. First, the aspect `hotel', which has a positive polarity and then the sub-aspect `workers'. We avoid the problem of deciding which is the opinion target by treating these as two separate opinions, whose targets are `hotel' and `workers'.

\begin{small}
\begin{exe}
\ex \label{implicit} \gll Igerilekua zegoen. \\
pool\smallsc{.abs.sg} was \\
\trans `There was a pool.'
\end{exe}
\end{small}

If there was an implicit opinion based on the presence or absence of a desirable aspect, such as the one seen in Example \ref{implicit}, we asked annotators to identify the phrase that indicates presence or absence, i.e. `there was', as the opinion phrase.

\begin{small}
\begin{exe}
\ex \label{span} \gll Langileek emandako arreta bikaina zen .\\
workers\smallsc{.erg.pl}   given\smallsc{.comp}    attention\smallsc{.abs.sg} excellent\smallsc{.abs.sg} was \\
\trans `The attention that the staff gave was excellent.'
\end{exe}
\end{small}

Finally, in order to improve overlap in span selection, we instructed annotators to choose the smallest span possible that retains the necessary information. Even after several iterations, however, there were still discrepancies with difficult examples, such as the one shown in Example \ref{span}, where the opinion target could be either `attention', `the attention', or `the attention that the staff gave'. 

\section{Benchmarks}

In order to provide a simple baseline, we frame the extraction of opinion holders, targets, and phrases as a sequence labeling task and map the NAF tags to BIO tags for the opinions in each review.  These tags serve as the gold labels which will need to be predicted at test time. We also perform classification of the polarity of opinion expressions. 

For the extraction of opinion holders, targets, and expressions we train a Conditional Random Field\footnote{We use the implementation available in \textit{sklearn\_crfsuite}.} (CRF) on standard features for supervised sequence labeling (word-, subword-, and part-of-speech information of the current word and previous words). For the classification of the polarity of opinion expressions, we use a Bag-of-Words approach to extract features and then train a linear SVM classifier\footnote{We use the liblinear implementation from \textit{sklearn}.}

For evaluation, we perform a 10-fold cross-validation with 80 percent of the data reserved for training during each fold. For extraction and classification, we report the weighted \F score. The results of the benchmark experiment (shown in Table \ref{table:benchmarks}) show that these simple baselines achieve results which are somewhat lower but still comparable to similar tasks in English \cite{Irsoy2014a}. The drop is not surprising given that we use a relatively simple baseline system and due to the fact that Catalan and Basque have richer morphological systems than English, which were not exploited.

\begin{table}[htp!]
\centering
\begin{tabular}{lcc}
\toprule
& \multicolumn{1}{c}{Catalan} & \multicolumn{1}{c}{Basque}\\
\cmidrule(lr){1-1}\cmidrule(lr){2-2}\cmidrule{3-3}
 \textit{Targets}   & .64  & .57 \\
 \textit{Expressions} & .52  & .54  \\
 \textit{Holders}     & .56   & .54 \\
 \textit{Polarity} & .80 & .84 \\
\bottomrule
\end{tabular}
\caption{Weighted \F scores for extraction of opinion targets, expressions and holders, as well as the weighted \F for classification of polarity.}
\label{table:benchmarks}
\end{table}

\section{Conclusion}
In this paper we have presented the MultiBooked corpus -- a corpus of hotel reviews annotated for aspect-level sentiment analysis available in Basque and Catalan. The aim of this annotation project is to allow researchers to enable research on supervised aspect-level sentiment analysis in Basque and Catalan, as well as provide useful data for cross- and multi-lingual sentiment analysis. We also provide inter-annotator agreement scores and benchmarks, as well as making the corpus available to the community.

\section{Bibliographical References}
\label{main:ref}

\bibliographystyle{lrec}
\bibliography{lit}

\section{Language Resource References}
\label{lr:ref}
\bibliographystylelanguageresource{lrec}
\bibliographylanguageresource{lit}

\end{document}